\documentclass{article}
\usepackage{spconf,amsmath,graphicx}

\title{A Weakly-Supervised Streaming Multilingual Speech Model with Truly Zero-Shot Capability}
\name{Jian Xue*, Peidong Wang*, Jinyu Li*, Eric Sun \thanks{* Equal Contribution}}
\address{Microsoft Speech Group, Redmond, WA, USA}
\begin{document}
%
\maketitle
\begin{abstract}
Neural transducers have been widely used in streaming automatic speech recognition (ASR) and speech translation (ST) tasks. In this paper, we introduce our work of building a \textbf{S}treaming \textbf{M}ultilingual \textbf{S}peech \textbf{M}odel ($SM^2$), which uses a single neural transducer model to transcribe or translate speech of multiple languages into texts of the target language without source language identification (LID). We adopt Transformer Transducer as the backbone of $SM^2$ since it has exhibited excellent streaming capability in previous works. Instead of using human labeled ST data, $SM^2$ is trained with weakly supervised data generated by converting the transcriptions in speech recognition corpora with a machine translation service. With 351 thousand hours of anonymized speech training data from 25 languages, $SM^2$ achieves comparable or even better ST quality than some recent popular large-scale non-streaming speech models. We further show that $SM^2$ has the truly zero-shot capability when expanding to new target languages, yielding high quality ST results for \{source-speech, target-text\} pairs that are not seen during training.
\end{abstract}
\section{Introduction}
\label{sec:intro}

With the advance of end-to-end (E2E) modeling \cite{E2EOverview}, E2E models emerge to dominate the fields of automatic speech recognition (ASR) \cite{watanabe2017hybrid, chiu2018state, he2019streaming, Li2020comparison} and speech translation (ST) \cite{Berard2016ST, vila2018end, pino2019harnessing, sperber2020speech}. 
This inspires many works of building a single E2E model for multilingual ASR \cite{watanabe2017language, toshniwal2018multilingual, zhou2022configurable} and multilingual ST \cite{inaguma2019multilingual, li2020multilingual}. The most popular E2E techniques for ASR are Connectionist Temporal Classification (CTC) \cite{graves2006connectionist}, Attention-based Encoder-Decoder (AED) \cite{cho2014learning}, and recurrent neural network Transducer (RNN-T) \cite{Graves-RNNSeqTransduction, prabhavalkar-comparison, Saon2021}. RNN-T does not have the conditional label independence assumption in CTC and also provides a more natural streaming solution than AED. Therefore, RNN-T has become the dominant E2E model for ASR tasks, especially in streaming applications. For E2E ST, most previous models are AED based because the attention mechanism in AED models can handle the word reordering challenge in ST. However, despite methods such as Monotonic Chunkwise Attention \cite{chiu2018monotonic}, Monotonic Infinite Lookback Attention \cite{Arivazhagan2019}, and Monotonic Multi-head Attention \cite{Ma2019Attention, Ma2021Attention} etc, AED models may not be a natural choice for streaming ST. In \cite{xue2022large}, Transformer Transducer (T-T) which uses streaming Transformer as the encoder of a neural transducer model, is shown to be a proper solution for streaming ST with high translation quality and low latency. In this work, we will also use T-T as the backbone model due to its high quality and low-latency properties. 

The most recent work for multilingual speech model is the Whisper model \cite{radfordrobust} which was trained with 680 thousand (K) hours of web data by carefully removing machine generated transcription. It is a Transformer AED model \cite{vaswani2017attention} which works in an offline mode and can perform many tasks including ASR, ST, spoken language identification (LID), and voice activity detection etc. The model obtains decent ASR and ST qualities when evaluated on tasks not observed during training, which was claimed as zero-shot capability in \cite{radfordrobust}. However, such capacity is usually considered as model robustness in prior works \cite{li2014overview}, and zero-shot translation is usually defined as the translation between language pairs whose data were never seen explicitly during model training \cite{johnson2017google}. Therefore an ST model with zero-shot translation capability should be trained without being exposed to the source-language audio and target-language text pairs.  

To build successful speech products in industry, there are many more practical factors need to be considered, such as streaming capability, inference cost, scalability of language expansion, and training data scarcity. Along this line of developing practical speech products, we introduce \textbf{S}treaming \textbf{M}ultilingual \textbf{S}peech \textbf{M}odel ($SM^2$), which can transcribe or translate multiple spoken languages into the transcription of a target language without source LID. $SM^2$ is different from \cite{radfordrobust} in the following major aspects:
\begin{enumerate}
    \item $SM^2$ is a streaming model which can be used in more applications. It also has much smaller model size, aligned with Green AI  \cite{schwartz2020green}.  
    \item $SM^2$ doesn't require source LID, thus can recognize and translate code-switch utterances with high quality.
    \item The ST training is totally weakly supervised without using any human labeled parallel corpus.
    \item $SM^2$ can be extended to additional target languages with a small amount of footprint increase. 
    \item $SM^2$ has the truly zero-shot ST capability. It can perform ST without being trained on the \{source-speech, target-text\} pairs.
\end{enumerate}

\section{Streaming Multilingual Speech Model}
\label{sec:SM2}

In this section we first introduce  $SM^2$ as a model initially designed to output texts in one target language only. Then we describe how to extend $SM^2$ with more output branches so that it can generate texts of multiple target languages. We will also discuss how that design enables $SM^2$ to perform zero-shot ST. 

\subsection{Streaming Multilingual Speech Model with Single Language Output}
\label{ssec:sm2single}
When we started to work on $SM^2$, our goal was to have a single streaming E2E speech model that can transcribe utterances of the target language (e.g., English)  and also translate multiple spoken languages (e.g., languages other than English) into the target language (e.g., English). So no matter which language the user speaks, the system will output the text in target language. Note that this is different from \cite{radfordrobust} which relies on user input to select between ASR and ST.
Another difference is that because of offline processing, \cite{radfordrobust} can first detect which language the user is speaking by looking at the whole utterance, and then use such LID information to guide ASR and ST. The LID information significantly boosts the quality of speech modeling \cite{toshniwal2018multilingual, kannan2019large}. However, streaming speech model cannot do this due to the latency constraint. Also if a system reply on LID information, it won't be able to process code-switch utterances properly.

\begin{figure}[t]
  \centering
  \includegraphics[width=1.0\linewidth]{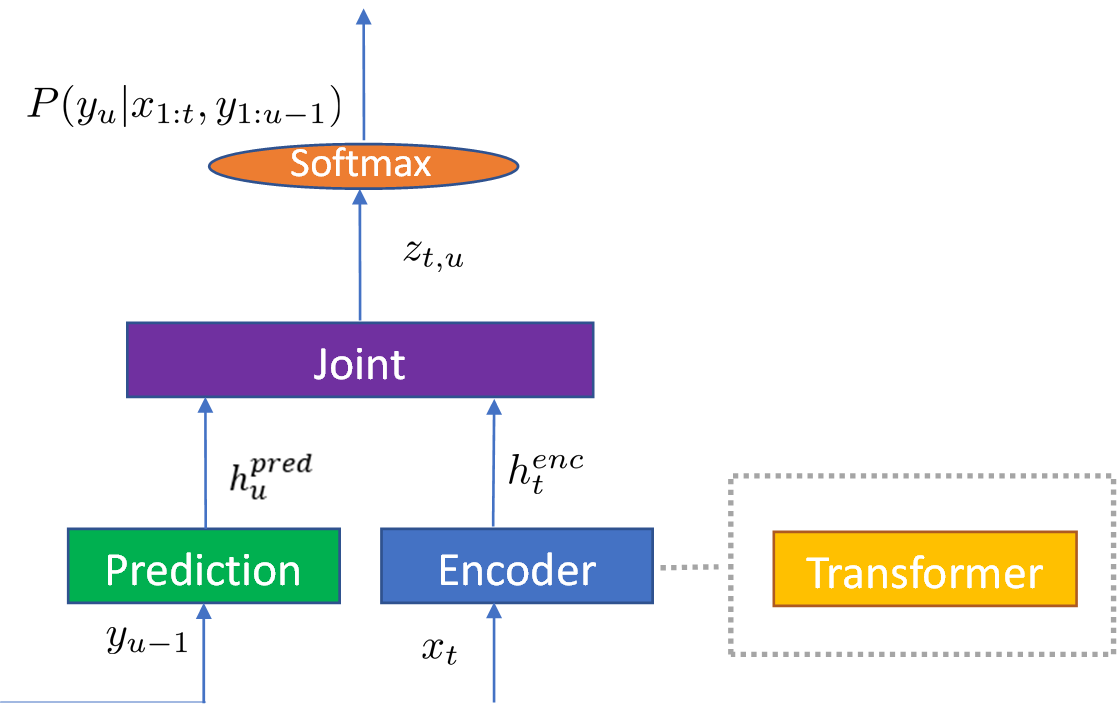}
  \caption{Illustration of a Transformer-Transducer}
  \label{fig:tt_module}
\end{figure}

The work in \cite{xue2022large} shows that neural transducer is a good solution for streaming ST with high translation quality and low latency. The reordering issue is handled naturally by a neural transducer since it dynamically decides read/write operations at each input feature frame. The neural Transducer has three components: an encoder network, a prediction network, and a joint network. When the encoder network is an RNN or a Transformer, the neural Transducer is called RNN-T or T-T, respectively. 
We build $SM^2$ with T-T which is shown in Fig. \ref{fig:tt_module}.  The encoder takes speech input $\textbf{x}_t$ to produce high-level speech representation $\textbf{h}_t^{enc}$ while the prediction network takes previous non-\textit{blank} output label $\textbf{y}_{u-1}$ from T-T to generate high-level representation $\textbf{h}_u^{pre}$. $t$ and $u$ denote the time and label steps, respectively. The joint network is a feedforward network which combines $\textbf{h}_t^{enc}$ and $\textbf{h}_u^{pre}$, and finally outputs the probability $P(\textbf{y}_u \in \textbf{Y} \cup \emptyset |\textbf{x}_{1:t}, \textbf{y}_{1:u-1})$, where $\textbf{Y}$ is the vocabulary list and $\emptyset$ denotes the \textit{blank} output.

\begin{figure}[t]
  \centering
  \includegraphics[width=1.0\linewidth]{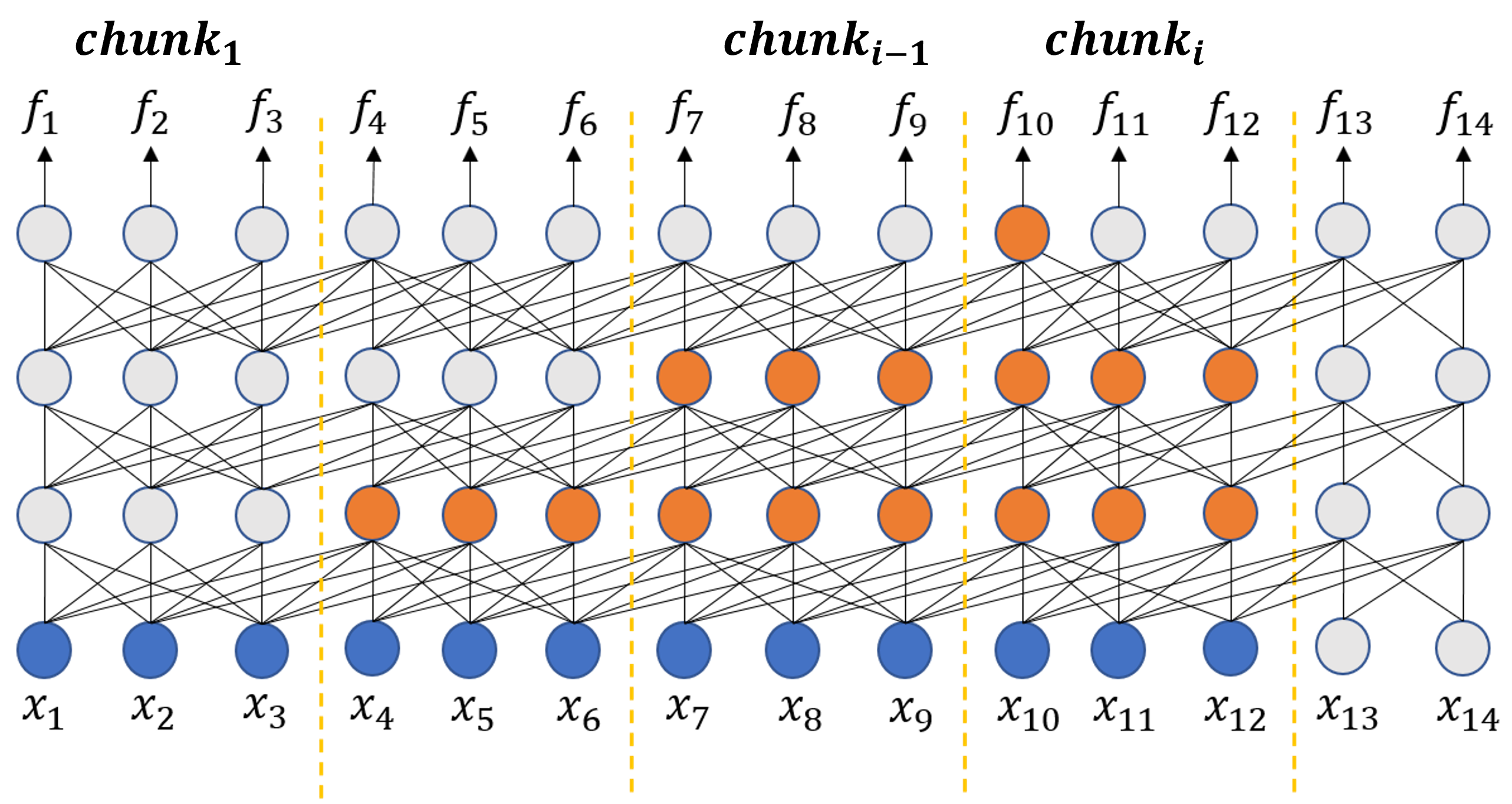}
  \caption{The reception field of a streaming T-T for generating output $f_{10}$. The chunk size is 3 and the number of left chunks is 1.}
  \label{fig:tt}
\end{figure}

We apply the attention mask proposed in \cite{xiechen2021tt} for the T-T to work in streaming mode. An example is shown in Fig. \ref{fig:tt}. We divide the speech inputs into chunks along time with chunk size $U$. Each frame can see fixed numbers of left chunks, and the left reception field increases linearly with the number of layers, enabling the model to use long history information for a better performance with much less computational cost than the model which uses full history at every layer. Within a chunk, all frame can see each other, but cannot see any frames in future chunks. Therefore, the algorithmic latency of such a T-T is the chunk size $U$.

In the experiment section, we will use chunk size $U$ to control the number of future speech frames that T-T can access. A larger chunk size gives better ASR and ST qualities since the model gets more information at each time step. 

When training $SM^2$, we pool the speech data of all languages together.
If the speech sample comes from the target language, it is an ASR task. Otherwise, it is an ST task. $SM^2$ does not need to be informed of whether the task is ASR or ST. Source LID is not needed either, so the system can process code-switch utterances naturally with high quality. Note that it is much more difficult to obtain a large-scale human labeled ST training set, as opposed to ASR. To solve this data scarcity issue, we use the weakly supervised method \cite{jia2019st} by calling a text based machine translation service to translate the ASR transcriptions to the target language. In this way, we do not use any human labeled ST data to train the model. 

\subsection{Language Expansion with Zero-Shot Capability}
\label{ssec:le}

\begin{figure}[t]
  \centering
  \includegraphics[width=1.0\linewidth]{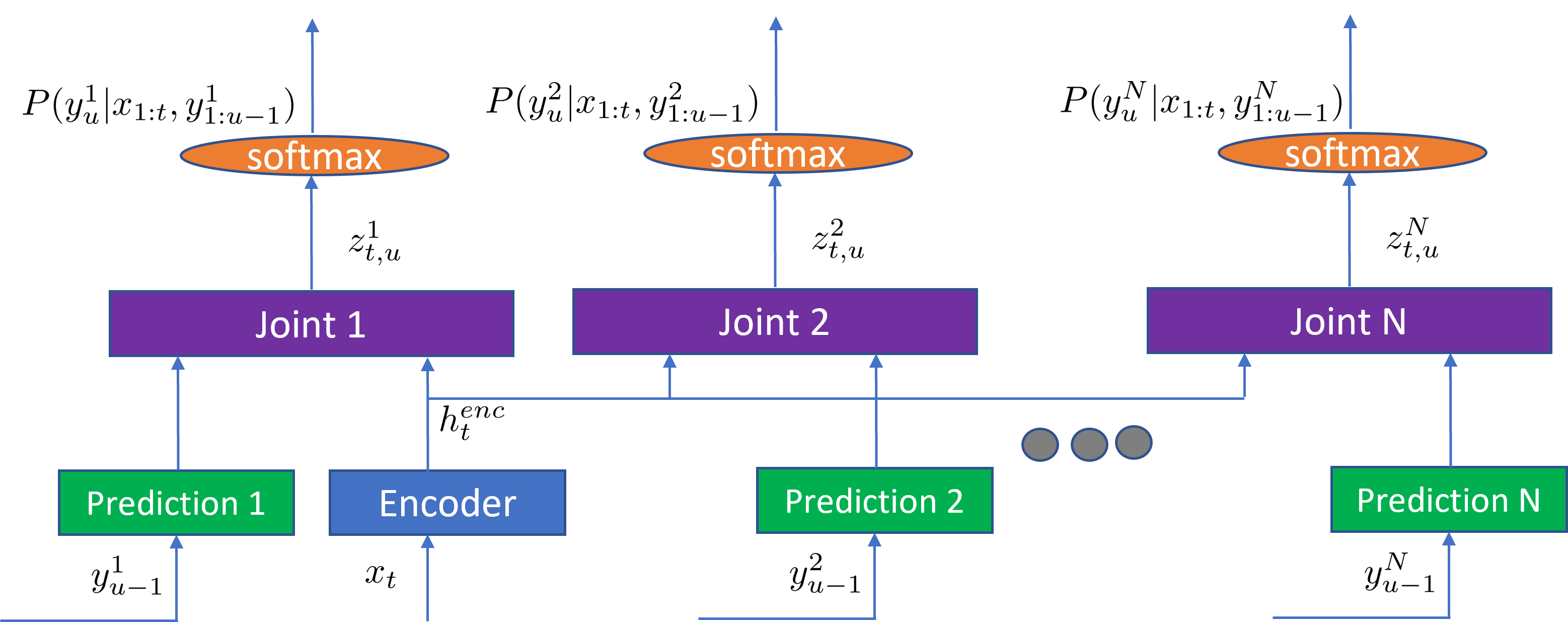}
  \caption{Illustration of language expansion.}
  \label{fig:expand}
\end{figure}

Scaling to more output languages is challenging to multilingual E2E ST models including $SM^2$. Suppose we have $S$ source languages and the target language is English, we only need to use $S$ language pairs to train a $SM^2$. However, if we want to support all $S$-language outputs, we need to have $S^2$ language pairs in the training set, introducing formidable training cost. Furthermore, after expanding to more output languages, we would like to avoid degrading the model performance on the original target language.

We propose a language expansion technique as shown in Fig. \ref{fig:expand}. We first train a $SM^2$  with one target language using the method described in Section \ref{ssec:sm2single}. When expanding to a new output language, we reuse and freeze the speech encoder from the previous model, and add new prediction and joint networks. 
Since prediction and joint networks have much less parameters compared with the encoder, the model size increase for adding a new target language is small.
The ST training data is again synthesized from the same ASR training corpus as what has been done for the first target language.

\begin{figure}[t]
  \centering
  \includegraphics[width=.8\linewidth]{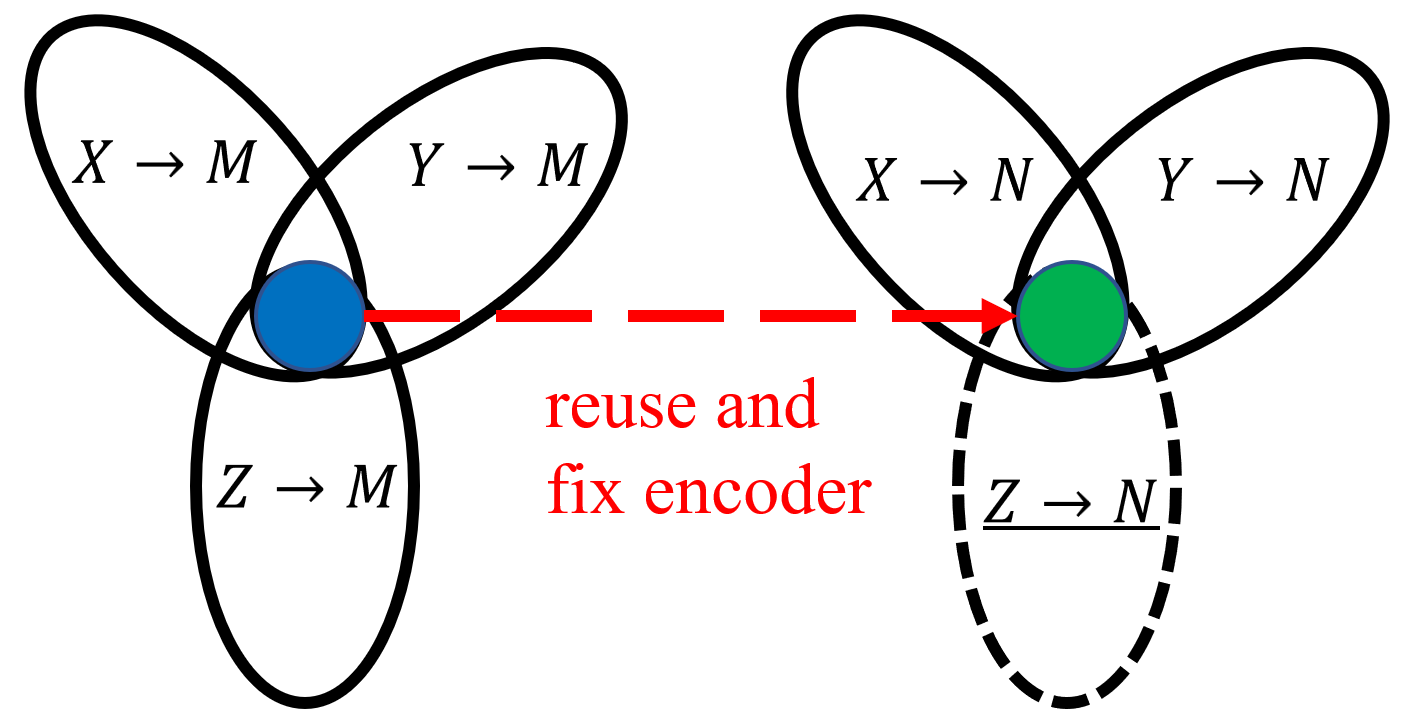}
  \caption{Illustration of the zero-shot mechanism. $Z \rightarrow N$ is not observed during training.}
  \label{fig:zero_shot}
\end{figure}

Our proposed method enables zero-shot ST, reducing the number of language pairs required during training, and thus drastically improve the training efficiency. 
Fig. \ref{fig:zero_shot} shows the mechanism facilitating the zero-shot capability of $SM^2$.
For a many-to-one $SM^2$ trained using $\{X, Y, Z\} \rightarrow M$ data where $X, Y, Z, M$ are different languages, we denote the shared representation space from the speech encoder as a blue circle, in which utterances in different languages have the same semantic meaning. Such inter-lingual space \cite{johnson2017google} can be obtained when we have a large amount of speech training data in multiple languages. For a new language output $N$, since we use the same multilingual ASR corpus to generate the transcriptions for $M$ and $N$  and we reuse and freeze the original speech encoder, its inter-lingual space represented by the green circle is the same as that of $\{X, Y, Z\} \rightarrow M$. Therefore, when we train the model for the new target language $N$ only with $\{X, Y\} \rightarrow N$ data, utterances in the inter-lingual space can also perform $Z \rightarrow N$ translation. Because of this calibration in the inter-lingual space and encoder freezing, $Z \rightarrow N$ translation can generalize to other utterances in language $Z$ shown in the dashed area in Fig. \ref{fig:zero_shot}, thus enables zero-shot translation.

\section{experiments}
\label{sec:typestyle}

To train $SM^2$, we use ASR training data from 25 languages: English (EN), Chinese (ZH), Portuguese (PT), Spanish (ES), Italian (IT), German (DE), French (FR), Japanese (JA), Russian (RU), Korean (KO), Polish (PL), Norwegian (NB), Hungarian (HU), Greek (EL), Czech (CS), Romanian (RO), Swedish (SV), Danish (DA), Finnish (FI), Dutch (NL), Slovenian (SL), Slovak (SK), Lithuanian (LT), Estonian (ET), and Bulgarian (BG). As shown in Table \ref{tab:language}, 
the corpora cover lower-, medium-, and high-resource languages containing [0.1K, 1K), [1K, 10K), and [10K, 100K] hours of training data, respectively.
The total number of training data is 351K hours. All the training  data is
anonymized with personally identifiable information removed.  A text based machine translation service is used to convert the ASR transcriptions into texts of the target language for ST training. 

\begin{table}[t]
    \centering
    \resizebox{0.96\columnwidth}{!}{
    \begin{tabular}{c|c}
    \hline
       hours &  languages \\
     \hline
      [0.1K, 1K)        &   SL, SK, LT, ET, BG   \\
      \hline
      [1K, 10K)       &       RU, KO, PL, NB, HU, EL, CS, RO, SV, DA, FI, NL  \\
    \hline
      [10K, 100K] &    EN, ZH, PT, ES, IT, DE, FR, JA    \\ 
     \hline
    \end{tabular}
    }
    \caption{Training data amount of 25 languages.}
    \vspace{-0.0cm}
    \label{tab:language}
\end{table}

We first trained several models based on the T-T structure described in Section \ref{ssec:sm2single} with same model structure but different algorithmic latencies (encoder lookahead). The above 25 languages are source input language and English is the target language. The encoder has 36 Transformer blocks, each contains 512 hidden nodes, 8 attention heads, and 4096 feedforward nodes. The prediction network has 2 LSTM layers with 1024 embedding dimension and 1024 hidden nodes. The joint network is a single feedforward layer with 512 nodes and the vocabulary size is 5K. The total number of parameters is 211 million (M). We investigated several chunk sizes as 0.32s, 1s, and 30s. We also trained another larger model with 30s chunk size, which has 24 Transformer blocks, each contains 1024 hidden nodes, 16 attention heads, and 4096 feedforward nodes. The total number of parameters for this model is 343M. The models with 30s chunk size are not feasible in a streaming system. We train such models as comparisons to see the up limit of the accuracy when we keep increasing the latency of the system. The 25 language to ZH model is based on the 211M model with 0.32s latency. The additional number of parameters added specifically for ZH output is 27M.

\subsection{Generating English Transcription from 25 Spoken Languages}

To compare the ST performance with the model in \cite{radfordrobust}, we take CoVoST 2 \cite{wang2021covost} as the benchmark and evaluate BLEU scores for both systems. The initial purpose of our $SM^2$ work is to build an in-house multilingual speech model, therefore we did not select the same language set as in CoVoST 2 and \textbf{did not include any CoVoST 2 data in training}.  We can only evaluate a subset of 12 language pairs that are observed in our training, as shown in Table \ref{tab:EN_BLEU}. The low-latency streaming $SM^2$ with 211M parameters and 0.32s chunk size 
has a BLEU score of 28.7 on average, much better than the small model in \cite{radfordrobust} which has 244M parameters and 30s chunk size\footnote{The models in \cite{radfordrobust} are offline models, but are operated in 30s chunks during inference.}. As we keep the model size but increase the chunk size, the $SM^2$ get better BLEU scores, 31.3 for the one with 1s chunk size, and 32.8 with 30s chunk size. Finally, increasing the number of parameters to 343M and the chunk size to 30s, the $SM^2$  reaches 33.7 BLEU score, slightly better than the largest model in \cite{radfordrobust}, which has 1550M parameters and 30s chunk size. 

\begin{table}[t]
    \centering
    \resizebox{0.96\columnwidth}{!}{
    \begin{tabular}{c|c|c|c|c|c|c}
    \hline
       & \multicolumn{2}{|c|}{Whisper \cite{radfordrobust}} &  \multicolumn{4}{|c}{$SM^2$} \\
     \hline
     model size & 244M &	1550M & \multicolumn{3}{|c|}{211M} & 343M \\
     \hline
     chunk size & 30s & 30s &	0.32s & 1s & 30s & 30s \\
     \hline

      DE$\rightarrow$EN        &   25.3	& 36.3 &	32.3  & 34.0 &	36.4 &	\textbf{37.8}       \\
      ZH$\rightarrow$EN	& 6.8 &	18.0 &	15.9  &  18.0 &	19.8 & \textbf{21.6}    \\
    JA$\rightarrow$EN	& 17.3	& \textbf{26.1}	& 20.1  &  21.6	& 23.5 & 25.4     \\
    RU$\rightarrow$EN	& 30.9	& 43.3	& 36.8  & 39.8 & 43.3 & \textbf{44.8}     \\
    NL$\rightarrow$EN	& 28.1	& 41.2	& 36.1  & 38.5	& 42.2 & \textbf{43.4}     \\
    ET$\rightarrow$EN	& 2.4	& 15.0	& 15.3  &  17.9 & 21.3 &\textbf{22.3}    \\
    SV$\rightarrow$EN	& 29.9	& \textbf{42.9}	& 33.6  &  37.1	& 36.5 & 33.8     \\
    SL$\rightarrow$EN	& 9.2	& 21.6	& 15.3  & \textbf{22.4} & 18.1 & 20.4     \\
    ES$\rightarrow$EN	& 33.0	& \textbf{40.1}	& 32.9  &  34.7 &	36.8 & 37.3    \\
    FR$\rightarrow$EN	& 27.3 	& \textbf{36.4}	& 31.5  & 33.0 &	34.9 & 35.9     \\
    IT$\rightarrow$EN	& 24.0	& 30.9	& 31.7  & 33.4 &	35.0 & \textbf{36.1}    \\
    PT$\rightarrow$EN	& 40.6	& \textbf{51.6}	& 42.4  &  44.7	& 45.6 &	45.8     \\
    \hline
    Average	& 22.9	&33.6	& 28.7  &  31.3	& 32.8 & \textbf{33.7}     \\
     \hline
    \end{tabular}}
    \caption{BLEU score comparison of different models on CoVoST 2 tasks with languages$\rightarrow$EN observed during training. The \textbf{bold} numbers indicate the best BLEU score for a specific language pair.}
    \vspace{-0.0cm}
    \label{tab:EN_BLEU}
\end{table}

Because \cite{radfordrobust} uses in-house training data, there is no apple-to-apple comparison between these models. However, we observe that 
\begin{itemize}
    \item State-of-the-art ST results can be achieved using weakly supervised ST training data, which is obtained by translating ASR transcriptions to texts of the target language with an MT system, without the need of any human labeled ST data.
    \item T-T based streaming multilingual ST models can yield very high translation quality even with a small model size and low latency, and without source LID information.
\end{itemize}

We compare different $SM^2$ variations in Table \ref{tab:ENASR} using our in-house ASR test set, which contains 1.8M words from various tasks. We also trained two ASR models as comparisons, with 0.32s chunk size and different model sizes, which can only transcribe English utterances. The 211M-parameter and 343M-parameter ASR models have the same T-T model structures as the $SM^2$ variations with the same model size, except that the chunk size may be different. 
For $SM^2$, both the 1s and 30s chunk size models are significantly better than the 0.32s model, showing the advantage of larger encoder lookahead. 
The ASR models with 0.32s chunk size outperform the corresponding $SM^2$ with the same chunk size in terms of WERs. This indicates that simply merging the transcriptions of ASR and ST together to train a single model is not optimal because the goal of ASR task is to precisely transcribe every word in the spoken utterance, whereas the goal of ST task is to convey the semantic meaning of an utterance. 

\begin{table}[t]
    \centering
    \resizebox{0.96\columnwidth}{!}{
    \begin{tabular}{c|c|c|c|c|c|c}
    \hline
    & \multicolumn{4}{|c|}{$SM^2$} &  \multicolumn{2}{|c}{ASR} \\
     \hline
       model size &  \multicolumn{3}{|c|}{211M} & 343M & 211M & 343M\\
     \hline
       chunk size &  0.32s & 1s & 30s & 30s & 0.32s & 0.32s \\
     \hline
        WER        &  8.81 &	8.18 &	7.55  & 7.27 & 7.72 & 7.36 \\
     \hline
    \end{tabular}
    }
    \caption{WERs of $SM^2$ and ASR models on 1.8M word test sets}
    \vspace{-0.0cm}
    \label{tab:ENASR}
\end{table}

\begin{table}[t]
  \centering
  \begin{tabular}{c|c|c|c|c|c}
    \hline
    \# source languages & 1 & 3 & 12 & 21 & 25 \\ \hline
      DE$\rightarrow$ZH & \textbf{2.2}    &   21.0  &  21.8 &22.5  &   21.3\\
      EN$\rightarrow$ZH	& \textbf{0.1}   & 28.9  & 29.2   & 29.3 & 28.2 \\
    JA$\rightarrow$ZH	&  \textbf{4.5}  & \textbf{11.4}  &    20.0   & 20.2 & 20.2 \\
    RU$\rightarrow$ZH	& \textbf{8.9}  &  \textbf{20.1 }&27.8      & 28.3 & 26.8 \\
    NL$\rightarrow$ZH	& \textbf{3.5}  & \textbf{18.4}  &    \textbf{22.6}  & 24.5 & 23.9 \\
    ET$\rightarrow$ZH	& \textbf{3.9}   & \textbf{9.7}  &  \textbf{12.4}     & 14.0 & 13.1 \\
    SV$\rightarrow$ZH	&  \textbf{5.8} & \textbf{19.3} & \textbf{22.4} & 23.4     & 23.1\\
    SL$\rightarrow$ZH	& \textbf{2.1}  & \textbf{6.3} &    \textbf{8.1}  & 8.5 & 8.7 \\
    ES$\rightarrow$ZH	& \textbf{2.0}  & \textbf{17.3}  & \textbf{22.3}  & \textbf{22.8}     & 25.0 \\
    FR$\rightarrow$ZH	& \textbf{2.9}  & \textbf{16.0} & \textbf{20.7}     & \textbf{21.7} & 23.8 \\
    IT$\rightarrow$ZH	& \textbf{2.3}  & \textbf{16.4}  &    \textbf{21.0}  & \textbf{22.2} & 24.2 \\
    PT$\rightarrow$ZH	&  \textbf{5.1}  & \textbf{21.6}  &    \textbf{26.4}  & \textbf{27.0} & 28.8 \\
    \hline
    Average & \textbf{3.6}	& 17.2  &   21.2 &    22.0 &  22.3 \\
\hline
    \end{tabular}
  \caption{BLEU score comparison among Chinese-output models trained with different numbers of source languages. The \textbf{bold} numbers indicate zero-shot evaluations, i.e., the \{source-speech, target-text\} pairs are not observed during training.}
  \label{tab:zero_shot}
\end{table}

\subsection{Language Expansion to Chinese with Zero-Shot Capability}
We evaluate the zero-shot capability when expanding the target language to ZH. We defined 5 training sets with different numbers of source languages as shown in Table \ref{tab:zero_shot}. 
The models were trained by reusing and freezing the encoder of the 25$\rightarrow$EN model which has 211M parameters and 0.32s latency. Then we train a new joint network and a new prediction network for ZH, which has the same structure as the 25$\rightarrow$EN model except that the vocabulary size is 15K.
The model in the $1$-source column was trained with only ZH speech data, and that in the $3$-source column used ZH, EN, and DE speech data. 
For the $12$-source column, the model was trained with ZH, EN, DE, CS, EL, HU, NB, PL, RO, RU, JA, and KO. The model in the $21$-source column used the speech from all languages except ES, FR, IT, and PT.
All these setups have missing \{source-speech, target-text\} pairs, indicated by the \textbf{bold} font in Table \ref{tab:zero_shot}. The language pairs used for training are selected randomly. We leave the investigation on language selection for zero-shot ST as future work. The model in the $25$-source column was trained with the speech from the full 25-language set.

As the number of source languages increases, the average BLEU scores keep improving. When the training data only has ZH speech, the ST quality is low, with an average BLEU score of 3.6.  In contrast, with only 3 source languages, $SM^2$ can already obtain 17.2 average BLEU score, close to the 22.3 score obtained using all 25 languages in training. When a half set of languages are observed during training (the $12$-source column), the resulting average BLEU score is 21.2,  only 1.1 away from the model trained with the full set of 25 languages. Note that in this $12$-source setup, 8 out 12 test language pairs are not observed during training.  Going from the $12$-source column to the $21$-source column and then the $25$-source column, we observed that new language pairs for training only give very limited BLEU score boosts from the zero-shot setups, e.g., 22.6 to 24.5 for NL$\rightarrow$ZH and 22.8 to 25.0 for ES$\rightarrow$ZH. This clearly demonstrates the zero-shot power of our models.

\subsection{Latency Measurement}
\label{sssec:latency}
We use average proportion (AP), average lagging (AL), and differentiable average lagging (DAL) proposed in \cite{ma2020simuleval} to measure the inference latencies of our $SM^2$ models. Table \ref{tab:latencies} describes the latency results, where all the numbers are averaged on all the CoVost2 sets used in Table \ref{tab:EN_BLEU}. Since the average audio length in the test set is around 5.7s, the models with 30s chunk size operate as offline model.

\begin{table}[t]
    \centering
    \begin{tabular}{c|c|c|c|c}
    \hline
       model size &  \multicolumn{3}{|c|}{211M} & 343M \\
     \hline
       chunk size &  0.32s & 1s & 30s & 30s \\
     \hline
        AP        &  0.69 &	0.76 &	1.0  & 1.0 \\
     \hline
        AL        &  1443 &	1870 &	5766  & 5766 \\
     \hline
        DAL        &  1423 &	1811 &	3458  & 3454 \\        
     \hline
    \end{tabular}
    \caption{Latency comparisons of $SM^2$ models on Covost2 sets, where AL and DAL values are in milliseconds (ms)}
    \vspace{-0.0cm}
    \label{tab:latencies}
\end{table}

\section{conclusions}
\label{sec:majhead}
In this paper, we presented our work of building a \textbf{S}treaming \textbf{M}ultilingual \textbf{S}peech \textbf{M}odel ($SM^2$) which is a single model for both ASR and ST without requiring task specifications from users.  We used Transformer Transducer as the backbone model for streaming capability and controlled the model latency by adjusting the chunk size of the speech encoder. The training data did not involve any human labeled ST sets. It was purely weakly supervised ST data generated by converting 351K hours of anonymized ASR data from 25 languages using text based machine translation service.
We designed a language expansion strategy which only adds a small amount of parameters to the original model and enables truly zero-shot capability for unseen \{source-speech, target-text\} pairs by leveraging interlingua representations.
For the task of generating English translations, the $SM^2$ with 0.32s algorithmic latency obtained much better BLEU score as the model with similar size (211M parameters vs. 244M parameters) in \cite{radfordrobust}, which is not streaming. The best $SM^2$ got similar BLEU score as the largest model in \cite{radfordrobust}, but model size is less than $1/4$ of that model. Finally, we demonstrated the strong zero-shot capability of $SM^2$ when expanding to support the Chinese output. The model trained with only half of language pairs is only 1.1 BLEU score behind the model trained with the full language pairs.  

From experiments, we noticed that simply merging ASR and ST texts together to train a single model may not be optimal due to the different goals of ASR and ST. In the future, we will explore better training methods to address this challenge and further advance $SM^2$.

\bibliographystyle{IEEEbib}
\bibliography{refs}

\begin{thebibliography}{10}

\bibitem{E2EOverview}
J.~Li,
\newblock ``Recent advances in end-to-end automatic speech recognition,''
\newblock {\em APSIPA Transactions on Signal and Information Processing}, vol.
  11, no. 1, 2022.

\bibitem{watanabe2017hybrid}
S.~Watanabe, T.~Hori, S.~Kim, J.~R. Hershey, and T.~Hayashi,
\newblock ``Hybrid {CTC}/attention architecture for end-to-end speech
  recognition,''
\newblock {\em IEEE Journal of Selected Topics in Signal Processing}, vol. 11,
  no. 8, pp. 1240--1253, 2017.

\bibitem{chiu2018state}
C.-C. Chiu, T.~N. Sainath, Y.~Wu, R.~Prabhavalkar, P.~Nguyen, Z.~Chen,
  A.~Kannan, et~al.,
\newblock ``State-of-the-art speech recognition with sequence-to-sequence
  models,''
\newblock in {\em Proceedings of ICASSP}, 2018, pp. 4774--4778.

\bibitem{he2019streaming}
Y.~He, T.~N. Sainath, R.~Prabhavalkar, I.~McGraw, R.~Alvarez, D.~Zhao,
  D.~Rybach, et~al.,
\newblock ``Streaming end-to-end speech recognition for mobile devices,''
\newblock in {\em Proceedings of ICASSP}, 2019, pp. 6381--6385.

\bibitem{Li2020comparison}
J.~Li, Y.~Wu, Y.~Gaur, C.~Wang, R.~Zhao, and S.~Liu,
\newblock ``On the comparison of popular end-to-end models for large scale
  speech recognition,''
\newblock in {\em Proceedings of Interspeech}, 2020, pp. 1--5.

\bibitem{Berard2016ST}
A.~Berard, O.~Pietquin, C.~Servan, and L.~Besacier,
\newblock ``Listen and translate: A proof of concept for end-to-end
  speech-to-text translation,''
\newblock in {\em NIPS Workshop on End-to-end Learning for Speech and Audio
  Processing}, 2016.

\bibitem{vila2018end}
L.~C. Vila, C.~Escolano, J.~A. Fonollosa, and M.~R. Costa-Jussa,
\newblock ``End-to-end speech translation with the transformer,''
\newblock in {\em Proceedings of Interspeech}, 2018, pp. 60--63.

\bibitem{pino2019harnessing}
J.~Pino, L.~Puzon, J.~Gu, X.~Ma, A.~D. McCarthy, and D.~Gopinath,
\newblock ``Harnessing indirect training data for end-to-end automatic speech
  translation: Tricks of the trade,''
\newblock {\em arXiv preprint arXiv:1909.06515}, 2019.

\bibitem{sperber2020speech}
M.~Sperber and M.~Paulik,
\newblock ``Speech translation and the end-to-end promise: Taking stock of
  where we are,''
\newblock in {\em Proceedings of the 58th Annual Meeting of the Association for
  Computational Linguistics}, 2020, pp. 7409--7421.

\bibitem{watanabe2017language}
S.~Watanabe, T.~Hori, and J.~R. Hershey,
\newblock ``Language independent end-to-end architecture for joint language
  identification and speech recognition,''
\newblock in {\em Proceedings of ASRU}, 2017, pp. 265--271.

\bibitem{toshniwal2018multilingual}
S.~Toshniwal, T.~N. Sainath, R.~J. Weiss, B.~Li, P.~Moreno, E.~Weinstein, and
  K.~Rao,
\newblock ``Multilingual speech recognition with a single end-to-end model,''
\newblock in {\em Proceedings of ICASSP}, 2018, pp. 4904--4908.

\bibitem{zhou2022configurable}
L.~Zhou, J.~Li, E.~Sun, and S.~Liu,
\newblock ``A configurable multilingual model is all you need to recognize all
  languages,''
\newblock in {\em Proceedings of ICASSP}, 2022, pp. 6422--6426.

\bibitem{inaguma2019multilingual}
H.~Inaguma, K.~Duh, T.~Kawahara, and S.~Watanabe,
\newblock ``Multilingual end-to-end speech translation,''
\newblock in {\em Proceedings of ASRU}, 2019, pp. 570--577.

\bibitem{li2020multilingual}
X.~Li, C.~Wang, Y.~Tang, C.~Tran, Y.~Tang, J.~Pino, A.~Baevski, A.~Conneau, and
  M.~Auli,
\newblock ``Multilingual speech translation with efficient finetuning of
  pretrained models,''
\newblock {\em arXiv preprint arXiv:2010.12829}, 2020.

\bibitem{graves2006connectionist}
A.~Graves, S.~Fern{\'a}ndez, F.~Gomez, and J.~Schmidhuber,
\newblock ``Connectionist temporal classification: labelling unsegmented
  sequence data with recurrent neural networks,''
\newblock in {\em Proceedings of the 23rd international conference on Machine
  learning}, 2006, pp. 369--376.

\bibitem{cho2014learning}
K.~Cho, B.~V. Merri{\"e}nboer, C.~Gulcehre, D.~Bahdanau, F.~Bougares,
  H.~Schwenk, and Y.~Bengio,
\newblock ``Learning phrase representations using {RNN} encoder-decoder for
  statistical machine translation,''
\newblock {\em arXiv preprint arXiv:1406.1078}, 2014.

\bibitem{Graves-RNNSeqTransduction}
A.~Graves,
\newblock ``Sequence transduction with recurrent neural networks,''
\newblock {\em arXiv preprint arXiv:1211.3711}, 2012.

\bibitem{prabhavalkar-comparison}
R.~Prabhavalkar, K.~Rao, T.~N. Sainath, B.~Li, L.~Johnson, and N.~Jaitly,
\newblock ``A comparison of sequence-to-sequence models for speech
  recognition,''
\newblock in {\em Proceedings of Interspeech}, 2017, pp. 939--943.

\bibitem{Saon2021}
G.~Saon, Z.~Tüske, D.~Bolanos, and B.~Kingsbury,
\newblock ``Advancing rnn transducer techology for speech recognition,''
\newblock in {\em Proceedings of ICASSP}, 2021, pp. 5654--5658.

\bibitem{chiu2018monotonic}
C.-C. Chiu and C.~Raffel,
\newblock ``Monotonic chunkwise attention,''
\newblock in {\em International Conference on Learning Representations}, 2018.

\bibitem{Arivazhagan2019}
N.~Arivazhagan, C.~Cherry, W.~Macherey, C.-C. Chiu, S.~Yavuz, R.~Pang, W.~Li,
  and C.~Raffel,
\newblock ``Monotonoic infinite lookback attention for simultaneous machine
  translation,''
\newblock in {\em Proceedings of the Annual Meeting of the Association for
  Computational Linguistics}, 2019, pp. 1313--1323.

\bibitem{Ma2019Attention}
X.~Ma, J.~Pino, J.~Cross, L.~Puzon, and J.~Gu,
\newblock ``Monotonic multihead attention,''
\newblock in {\em Proceedings of International COnference on Learning
  Representations}, 2019.

\bibitem{Ma2021Attention}
X.~Ma, Y.~Wang, M.~J. Dousti, P.~Koehn, and J.~Pino,
\newblock ``Streaming simultaneous speech translation with augmented memory
  transformer,''
\newblock in {\em Proceedings of ICASSP}, 2021, pp. 7523--7527.

\bibitem{xue2022large}
J.~Xue, P.~Wang, J.~Li, M.~Post, and Y.~Gaur,
\newblock ``Large-scale streaming end-to-end speech translation with neural
  transducers,''
\newblock in {\em Proceedings of Interspeech}, 2022, pp. 3263--3267.

\bibitem{radfordrobust}
A.~Radford, J.~W. Kim, T.~Xu, G.~Brockman, C.~McLeavey, and I.~Sutskever,
\newblock ``Robust speech recognition via large-scale weak supervision,''
\newblock 2022.

\bibitem{vaswani2017attention}
A.~Vaswani, N.~Shazeer, N.~Parmar, J.~Uszkoreit, L.~Jones, A.~N. Gomez,
  {\L}.~Kaiser, and I.~Polosukhin,
\newblock ``Attention is all you need,''
\newblock in {\em Advances in Neural Information Processing Systems}, 2017, pp.
  6000--6010.

\bibitem{li2014overview}
J.~Li, L.~Deng, Y.~Gong, and R.~Haeb-Umbach,
\newblock ``An overview of noise-robust automatic speech recognition,''
\newblock {\em IEEE/ACM Transactions on Audio, Speech, and Language
  Processing}, vol. 22, no. 4, pp. 745--777, 2014.

\bibitem{johnson2017google}
M.~Johnson, M.~Schuster, Q.~V. Le, et~al.,
\newblock ``Google's multilingual neural machine translation system: Enabling
  zero-shot translation,''
\newblock {\em Transactions of the Association for Computational Linguistics},
  vol. 5, pp. 339--351, 2017.

\bibitem{schwartz2020green}
R.~Schwartz, J.~Dodge, N.~A. Smith, and O.~Etzioni,
\newblock ``Green {AI},''
\newblock {\em Communications of the ACM}, vol. 63, no. 12, pp. 54--63, 2020.

\bibitem{kannan2019large}
A.~Kannan, A.~Datta, T.~N. Sainath, E.~Weinstein, B.~Ramabhadran, Y.~Wu,
  A.~Bapna, Z.~Chen, and S.~Lee,
\newblock ``Large-scale multilingual speech recognition with a streaming
  end-to-end model,''
\newblock in {\em Proceedings of Interspeech}, 2019, pp. 2130--2134.

\bibitem{xiechen2021tt}
X.~Chen, Y.~Wu, Z.~Wang, S.~Liu, and J.~Li,
\newblock ``Developing real-time streaming transformer transducer for speech
  recognition on large-scale dataset,''
\newblock in {\em Proceedings of ICASSP}, 2021, pp. 5904--5908.

\bibitem{jia2019st}
Y.~Jia, M.~Johnson, W.~Macherey, R.~J. Weiss, Y.~Cao, C.-C. Chiu, N.~Ari,
  S.~Laurenzo, and Y.~Wu,
\newblock ``Leveraging weakly supervised data to improve end-to-end
  speech-to-text translation,''
\newblock in {\em Proceedings of ICASSP}, 2019, pp. 7180--7184.

\bibitem{wang2021covost}
C.~Wang, A.~Wu, J.~Gu, and J.~Pino,
\newblock ``{CoVoST} 2 and massively multilingual speech translation,''
\newblock in {\em Proceedings of Interspeech}, 2021, pp. 2247--2251.

\bibitem{ma2020simuleval}
X.~Ma, M.~J. Dousti, C.~Wang, J.~Gu, and J.~Pino,
\newblock ``{SIMULEVAL}: An evaluation toolkit for simultaneous translation,''
\newblock in {\em Proceedings of the Conference on Empirical Methods in Natural
  Language Processing: System Demonstrations}, 2020, pp. 144--150.

\end{thebibliography}

\end{document}